%% file: main.tex
\documentclass[journal]{IEEEtran}
\IEEEoverridecommandlockouts

\input{component_structure_sections/packages}

\graphicspath{ {./images/} } 

\def\BibTeX{{\rm B\kern-.05em{\sc i\kern-.025em b}\kern-.08em
    T\kern-.1667em\lower.7ex\hbox{E}\kern-.125emX}}

\input{component_structure_sections/header}

\renewcommand\thesection{\arabic{section}}
\renewcommand\thesubsection{\thesection.\arabic{subsection}}
\renewcommand\thesubsubsection{\thesubsection.\arabic{subsubsection}}

\renewcommand\thesectiondis{\arabic{section}}
\renewcommand\thesubsectiondis{\thesectiondis.\arabic{subsection}}
\renewcommand\thesubsubsectiondis{\thesubsectiondis.\arabic{subsubsection}}

\makeatletter
\def\subsubsection{%
  \@startsection
    {subsubsection}                 
    {3}                             
    {\parindent}                    
    {3.5ex plus 1.5ex minus 1.5ex}  
    {0.7ex plus .5ex minus 0ex}     
    {\normalfont\normalsize\itshape}
}
\makeatother

\begin{document}

\maketitle


\input{component_subject_sections/Abstract/abstract}




\section{Introduction}\label{sect::intro}
\input{component_subject_sections/Introduction/introduction}



\section{Background}\label{sect::back}
\input{component_subject_sections/Background/background}



\section{Prior Art}\label{sect::prior_art}
\input{component_subject_sections/Prior_art/prior_art}



\section{Methodology}\label{sect::core}
\input{component_subject_sections/Methodology/methodology}



\section{Results And Discussion}\label{sect::results}
\input{component_subject_sections/Results/results}






\section{Conclusion And Future Work}\label{sect::conc}
\input{component_subject_sections/Conclusion/conclusion}






\bibliographystyle{ieeetr} 
\bibliography{ref} 


\end{document}

%% file: component_structure_sections/packages.tex
\usepackage{cite}
\usepackage{amsmath,amssymb,amsfonts}
\usepackage{algorithm}
\usepackage{algpseudocode}
\usepackage{textcomp}
\usepackage{xcolor}
\usepackage{soul}
\usepackage{graphicx} 
\usepackage{comment}
\usepackage{pdfpages}
\usepackage{xurl}
\usepackage{etoolbox}
\usepackage{hyperref}
\usepackage{csquotes}

\usepackage{tabularx}
\usepackage[most]{tcolorbox}
\usepackage{caption} 
\usepackage{subcaption}
\usepackage{array}
\usepackage{color}
\usepackage{diagbox}
\usepackage{verbatim}
\usepackage{multicol}
\usepackage{multirow}
\usepackage{gensymb}

\input{component_structure_sections/vars}

\makeatletter
\newcommand{\linebreakand}{
  \end{@IEEEauthorhalign}
  \hfill\mbox{}\par
  \mbox{}\hfill\begin{@IEEEauthorhalign}
}
\makeatother

%% file: component_structure_sections/vars.tex
\newbool{author_sep}
\boolfalse{author_sep}

%% file: component_structure_sections/header.tex
\title{RAAD-LLM: Adaptive Anomaly Detection Using LLMs and RAG Integration\\
\thanks{This material is based upon work supported by the Engineering Research and Development Center - Information Technology Laboratory (ERDC-ITL) under Contract No. W912HZ23C0013. Any opinions, findings and conclusions or recommendations expressed in this material are those of the author(s) and do not necessarily reflect the views of the ERDC-ITL.}
}
\ifbool{author_sep}
{
}
{
    \author{
        \IEEEauthorblockN{
            Alicia Russell-Gilbert\IEEEauthorrefmark{1}\IEEEauthorrefmark{3},
            Sudip Mittal\IEEEauthorrefmark{1},
            Shahram Rahimi\IEEEauthorrefmark{1},\\
            Maria Seale\IEEEauthorrefmark{2},
            Joseph Jabour\IEEEauthorrefmark{2},
            Thomas Arnold\IEEEauthorrefmark{2},
            Joshua Church\IEEEauthorrefmark{2}\\
        }
        \IEEEauthorblockA{
            \IEEEauthorrefmark{3} corresponding author
            \IEEEauthorrefmark{1}
            \textit{Computer Science \& Engineering Department at}
            \textit{Mississippi State University} \\
            \{ar2836\}@msstate.edu, \{mittal, rahimi\}@cse.msstate.edu \\
        }
        \IEEEauthorblockA{
            \IEEEauthorrefmark{2}
            \textit{Engineer Research and Development Center at the} 
            \emph{Department of Defense}
            \\\{maria.a.seale, joseph.e.jabour, thomas.l.arnold, joshua.q.church\}@erdc.dren.mil
        }
    }
}

%% file: component_subject_sections/Abstract/abstract.tex
\begin{abstract}
Anomaly detection in complex industrial environments poses unique challenges, particularly in contexts characterized by data sparsity and evolving operational conditions. Predictive maintenance (PdM) in such settings demands methodologies that are adaptive, transferable, and capable of integrating domain-specific knowledge. In this paper, we present RAAD-LLM, a novel framework for adaptive anomaly detection, leveraging large language models (LLMs) integrated with Retrieval-Augmented Generation (RAG). This approach addresses the aforementioned PdM challenges. By effectively utilizing domain-specific knowledge, RAAD-LLM enhances the detection of anomalies in time series data without requiring fine-tuning on specific datasets. The framework's adaptability mechanism enables it to adjust its understanding of normal operating conditions dynamically, thus increasing detection accuracy. We validate this methodology through a real-world application for a plastics manufacturing plant and the Skoltech Anomaly Benchmark (SKAB). Results show significant improvements over our previous model with an accuracy increase from $70.7\%$ to $88.6\%$ on the real-world dataset. By allowing for the enriching of input series data with semantics, RAAD-LLM incorporates multimodal capabilities that facilitate more collaborative decision-making between the model and plant operators. Overall, our findings support RAAD-LLM's ability to revolutionize anomaly detection methodologies in PdM, potentially leading to a paradigm shift in how anomaly detection is implemented across various industries.
\\
\end{abstract}

%% file: component_subject_sections/Introduction/introduction.tex
In the rapidly evolving landscape of AI and knowledge-based systems, expert systems have emerged as powerful tools for incorporating domain expertise and specialized knowledge into models. Furthermore, they continue to be applied across domains such as engineering, agriculture, and manufacturing \cite{Blagoveshchenskiy_2020, ABIOYE2020105441, FILTER201561, shahzadi2016internet}. These systems aim to emulate the decision-making capabilities of human experts and they offer the potential to improve the performance of other approaches in many ways. 

In particular, domain knowledge integration helps identify relevant features and patterns that might be missed by purely data-driven approaches. In addition, expert-guided rules and thresholds enable more accurate anomaly detection by incorporating industry-specific maintenance criteria. Lastly, expert knowledge supports more robust fault detection by considering equipment-specific degradation patterns and maintenance history. These performance improvements lead to more accurate and reliable predictive maintenance models that better reflect real-world operational conditions. However, a persistent challenge is the gap between expert systems and the domain experts whose knowledge they aim to capture and apply.

This gap manifests itself in multiple ways such as communication barriers between AI developers and subject matter experts, difficulties in accurately translating complex human expertise into computational models, and resistance from experts who may view such systems as threats rather than aids. These issues can have significant consequences, leading to expert systems that fail to capture the nuanced decision-making processes of human experts, are difficult to update and maintain, or face limited adoption in real-world settings.

Bridging this gap is crucial because it can lead to more accurate and comprehensive expert systems that truly reflect the depth and breadth of human expertise. In addition, it can facilitate more effective knowledge transfer and preservation. Lastly, it can promote greater acceptance and integration of expert systems in professional practice. This integration has the potential to revolutionize fields such as healthcare, network security, environmental sciences, and manufacturing.

While expert systems can be applied across various domains, one critical area where they are particularly valuable is in maintaining complex engineered systems. Engineered systems that are vital to our daily operations degrade over time and can fail. These failures often lead to consequences that range from minor inconveniences to catastrophic events. To prevent such failures, maintenance practices such as condition-based maintenance (CBM) and predictive maintenance (PdM) are used. While CBM involves performing maintenance based on system conditions, PdM enhances this approach by using machine learning (ML) to make more proactive and targeted decision-making. This research focuses on developing an expert system for PdM that integrates domain expertise to enhance model performance and bridge the gap between automated systems and human experts in real-world applications.

PdM is challenging under real-world conditions as a result of non-stationary sensor data. Factors such as varying operational settings and individual machine deterioration are common causes of non-stationary sensor readings \cite{saurav2018online}. This heterogeneity in the relationship between operational data and system health requires regular updates of the normative profile used for the identification of degradation \cite{Steenwinckel2021FLAGSAM, Steenwinckel2018AdaptiveAD}. To address these challenges, an adaptive approach rather than traditional PdM methods should be employed. This would allow for better accommodation of shifts in sensor data characteristics while maintaining high fault detection accuracy.

Unique production systems and domain constraints require tailored PdM approaches across industries. Integrating expert knowledge enables robust domain-specific implementations. Yet, this knowledge often limits the applicability across domains. Therefore, retraining or fine-tuning on the applied dataset with related domain-specific knowledge would typically be required. However, event data needed to fine-tune or retrain may be scarce \cite{XIA2018255}. This is because some critical assets are not allowed to run to failure. Therefore, ideally, PdM models should be transferable in data-sparse scenarios. 

Transferable models that excel in \textquote{few-shot} and \textquote{zero-shot} scenarios can perform well on limited training data across diverse systems and domains. Recent work suggests that pretrained large language models (LLMs) offer notable few/zero-shot capabilities and transferability \cite{LLM_few_shot, LLM_few_shot_health, LLM_few_shot_prog}. The extension of LLMs beyond natural language to the time series domain showcases their broader potential \cite{LLM_in_TS, LLM_zero_shot_TS}. Repurposing pretrained LLMs for the PdM use-case can improve the transferability of other approaches in data-constrained environments. 

In light of the given challenges, PdM represents a particularly difficult application area of expert systems where domain expertise is crucial. These challenges underscore that for data-constrained, complex and dynamic industrial environments; there is a critical need for adaptable and transferable methodologies to enhance anomaly detection and therefore, prevent costs associated with system failures. Furthermore, multimodal strategies would more easily allow for the enriching of input series data with domain-specific knowledge. Consequently, expert systems would more accurately translate complex subject matter expertise into its computational models, be easier to update and maintain, and be more accepted in real-world settings.

This paper examines the application of RAAD-LLM (AAD-LLM with RAG integration), a novel expert system for anomaly detection in PdM scenarios that builds off of our previous work titled \textquote{AAD-LLM: Adaptive Anomaly Detection Using Large Language Models} \cite{10825679}. Specifically, this framework utilizes pretrained LLMs for anomaly detection in complex and data-sparse manufacturing systems. The proposed methodology does not require any training or fine-tuning on the dataset it is applied to. In addition, the architecture overcomes the issue of concept drift in dynamic industrial settings by integrating an adaptability mechanism. Furthermore, the framework is multimodal; thereby enabling more collaborative decision-making between the expert system and plant operators by allowing for the enriching of input time series data with semantics. Therefore, RAAD-LLM is shown to be a robust, transferable, and more widely adoptable expert system that supports rather than replaces human expertise.

The main contributions of this work are as follows:
\begin{itemize}
    \item We present a novel anomaly detection framework (RAAD-LLM) and explore the integration of a Retrieval-Augmented Generation (RAG) pipeline into the AAD-LLM architecture to improve its performance.
    \item We show that by leveraging pretrained LLMs, RAAD-LLM is transferable with zero-shot capabilities in comparison to other anomaly detection methodologies.
    \item RAAD-LLM is shown to be effective by applying it to a real-world use-case at a plastics manufacturing plant.
    \item We show that the adaptability mechanism of RAAD-LLM enables the model to adjust to evolving conditions, consequently enhancing detection accuracy.
    \item RAAD-LLM is shown to be multimodal; thereby delivering more context-aware detection to enable robust, domain-specific implementations in collaboration with plant operators.
\end{itemize}

The remaining sections of this paper are as follows. Section \ref{sect::back} discusses the background and foundational work for our proposed methodology. Section \ref{sect::prior_art} examines the state-of-the-art in LLM time series tasks and adaptive anomaly detection methods. Section \ref{sect::core} provides insight on the RAAD-LLM architecture and methodology. Section \ref{sect::results} explains evaluation results and implications of findings. Finally, Section \ref{sect::conc} concludes the paper and discusses limitations for future work.

%% file: component_subject_sections/Background/background.tex
\begin{figure*}
    \includegraphics[width=0.9\textwidth]{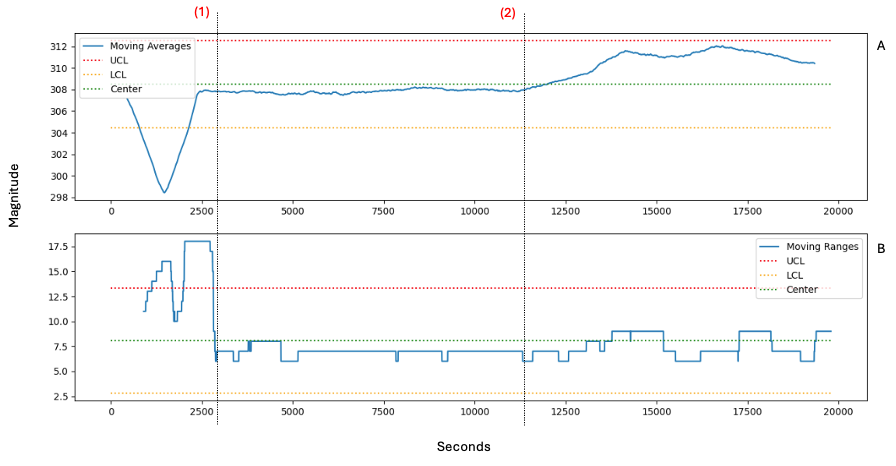}
    \caption{SPC technique of MAMR to set control limits for process stability in a query series $Q_i$. Figure A and Figure B are moving average and moving range, respectively. UCL is the defined upper control limit and LCL is the defined lower control limit. Series data points outside of control limits are deemed \textquote{out of statistical control} and are labeled as anomalous. Out of control points can be seen before line \textcolor{red}{(1)}. Points between lines \textcolor{red}{(1)} and \textcolor{red}{(2)} represent a stable process.  Points after line \textcolor{red}{(2)} also represent a stable process, however, they are trending towards out of control. These points, therefore, are potentially problematic. RAAD-LLM is applied to all points within control limits to enhance anomaly detection.}
    \label{fig:fig_meth3}
\end{figure*}

This section serves as a background for understanding LLMs and adaptive anomaly detection as presented in this paper. It aims to provide key terms, baseline definitions, and relevant mathematical notations that are essential for comprehending the concepts discussed. Additionally, this section briefly discusses the initial stages of our research endeavor. It describes the preliminary investigations conducted to lay the groundwork for our current work.

\subsection{Fundamental Concepts and Terminology}

A \textbf{large language model (LLM)} is trained on sequences of tokens and encodes an auto-regressive distribution, where the probability of each token depends on the preceding ones \cite{LLM_zero_shot_TS}. More simply, an LLM is trained on sequences of words or word pieces, and the output is the likelihood of the next word in a sequence given the previous words (i.e., context-aware embeddings). Each model includes a tokenizer that converts input strings into token sequences. Models like GPT-3 and LLaMA-2 can perform zero-shot generalization, effectively handling tasks without specific training  \cite{LLM_zero_shot_TS}. For this work, we repurpose an LLM for time series anomaly detection while keeping the backbone language model intact \cite{Time_LLM}. A binarization function is then applied to the outputs of the LLM to map them to $\{0,1\}$ to obtain the final predictions. The exact binarization function is use-case specific.

\textbf{Transfer learning} is a ML technique where the knowledge gained through one task is applied to a related task  with low/no retraining \cite{10.5555/1803899}. Specifically, in transfer learning, we train a model to perform a specific task on the source domain and then make certain modifications to give us good predictions for a related task on the target domain where data is (usually) scarce or a fast training is needed \cite{Neyshabur2020WhatIB}. For this work, we leverage a pretrained LLMs' text synthesizing and reasoning abilities acquired through training on a source domain by transferring this task knowledge to our PdM use-case. Specifically, we show that pretrained LLMs can effectively predict anomalies in time series data by transferring its text synthesizing and reasoning knowledge to our target manufacturing domain. 

\textbf{Concept drift} is the phenomenon where the statistical properties of a domain changes over time, which can then result in a deterioration of models that have previously been trained within that domain \cite{Stachl2019RepositoryPR, BERGGREN2024112465}. In particular, it can lead to a degradation of performance of static models as they become less effective in detecting anomalies.  For example, in manufacturing, the statistical properties of raw material attributes change over time. Therefore, if these variables are used as product quality predictors, the resulting models may decrease in validity. 

\textbf{Adaptive anomaly detection (AAD)} encompasses techniques that can detect anomalies in data streams or in situations where concept drift is present. These techniques make models capable of automatically adjusting their detection behavior to changing conditions in the deployment environment or system configuration while still accurately recognizing anomalies \cite{Steenwinckel2021FLAGSAM, Steenwinckel2018AdaptiveAD}. For this work, the \textbf{adaptability mechanism} refers to the feature that enables the model's definition of normality and related statistical measures to adjust with each new data instance. 

\textbf{Windowing} refers to dividing a time series into smaller, manageable segments, which are then processed individually. Windowing (or \textit{sliding window technique}) is used extensively for anomaly detection in time series data due to its many benefits \cite{w13131862}. For our use-case, dividing the time series into windows helps to preserve local information that might be lost when considering the entire time series as a whole and reduce computational load since models can handle smaller inputs more efficiently.  

A process is said to be \textquote{in statistical control} if it is not experiencing out of control signals or significant variations beyond normal statistical variations \cite{mcshane2023asq}. \textbf{Statistical process control (SPC)} techniques are commonly used in manufacturing for monitoring sequential processes (e.g., production lines) to make sure that they work stably and satisfactorily \cite{Qiu2018SomePO}. In monitoring the stability of a process, SPC plays an essential role \cite{qiu2013introduction, SONG2023109469}. The idea is that processes that are in statistical control are deemed to be \textit{stable} processes \cite{mcshane2023asq}. For this work, stable processes form a baseline for \textbf{normal} process behavior. The selection of SPC techniques are use-case specific. For this work, moving average moving range (MAMR) is implemented.

The univariate \textbf{MAMR} charts are plotted for each process variable in a time-series instance as shown in \autoref{fig:fig_meth3}. This aspect of plotting and analyzing the MAMR charts for all process variables in parallel will cause an increase in the Type I error rate. Upper (UCL) and lower (LCL) control limits for the moving average ($X$) and moving range ($mR$) charts are calculated as follows.

\setlength\parindent{50pt} $X$ Chart:
\begin{equation} \label{eq:xucl}
UCL = \overline{X} + 2.66\overline{R}
\end{equation}

\begin{equation} \label{eq:xlcl}
LCL = \overline{X} - 2.66\overline{R}
\end{equation}

$mR$ Chart:
\begin{equation} \label{eq:rucl}
UCL= 3.27\overline{R}
\end{equation}

\setlength\parindent{14pt} 
The values $2.66$ and $3.27$ are often used as multipliers for estimating control limits in the MAMR chart. However, these multipliers can significantly widen the control limits, making them less sensitive to minor shifts or variations in the process. Therefore, it is important to analyze historical data to determine the typical variability in the process under consideration and select multipliers that reflect the process's actual behavior while maintaining sensitivity.


\subsection{Investigated Approaches in Expert Systems for PdM}

The need for effective methodologies in predictive maintenance (PdM) is critical in complex and evolving industrial environments. In prior research, we explored the challenges inherent in conventional PdM approaches, particularly emphasizing their limitations in transferability across varied operational contexts and their lack of multimodality. Our foundational work, AAD-LLM, leveraged the capabilities of LLMs to establish a novel framework for anomaly detection in manufacturing settings characterized by sparse data.

In the development of AAD-LLM, we focused on the inherent strengths of pretrained LLMs and their capacity for zero-shot learning, which does not require extensive retraining on domain-specific datasets. The model was designed to convert anomaly detection into a language-based task by enriching time series data with semantic context derived from domain knowledge. Results from our implementation on real-world data (shown in \autoref{table:tab_results1}) demonstrated an accuracy of $70.7\%$. Evaluation metrics showed the model's potential in detecting anomalies effectively, even in data-constrained scenarios. However, we recognized that the model's performance in making comparative evaluations between the historical normal and the observed statistics was inconsistent, pointing to the necessity for a more robust computational mechanism.

\subsection{RAG}

RAG stands for Retrieval-Augmented Generation, a technique that enhances LLMs by integrating external, reliable, and up-to-date knowledge during the generation process \cite{Fan2024ASO}.  Specifically, RAG first invokes the retriever to search and extract the relevant documents from external databases, which are then leveraged as the context to enhance the generation process \cite{Fan2024ASO, Izacard2020LeveragingPR}. In practice, RAG requires minimal or even no additional training \cite{Fan2024ASO, Ram2023InContextRL}. 

The RAG approach has been shown to improve the baseline performance of LLMs.  For example, RAG has been shown to improve the performance of the question and answering task \cite{Khattab2021BaleenRM}. In another paper, Melz \cite{Melz2023EnhancingLI} showed that RAG improves the problem-solving abilities of LLMs. In addition to these works, a comprehensive review paper examined various RAG paradigms and emphasized RAG's significant advancement in enhancing the capabilities of LLMs \cite{Gao2023RetrievalAugmentedGF}.

Interim results for the AAD-LLM framework revealed that LLMs hold considerable promise in anomaly detection tasks for the PdM use-case. In addition to enhancing anomaly detection through the repurposing of LLMs, the current work introduces RAG. We hypothesize that integrating RAG into our existing framework would improve its performance. By facilitating the retrieval of relevant data for mathematical comparisons, RAG could enhance both the accuracy and applicability of AAD-LLM in industrial settings, where domain expertise is critical for interpreting complex scenarios. Thus, this work seeks to expand upon the insights gained from our earlier research, potentially leading to a paradigm shift in how anomaly detection is implemented across various industries.

%% file: component_subject_sections/Prior_art/prior_art.tex

This section examines recent advancements in applying LLMs to time series tasks, including forecasting, classification, anomaly detection, and imputation. It highlights the strengths and weaknesses of state-of-the-art methods. Additionally, it reviews prior research in AAD techniques that combine semantics with ML.

\subsection{LLMs for Time Series Tasks}
Traditional analytical methods that rely on statistical models and deep learning methods based on recurrent neural networks (RNNs) have dominated the domain of time series forecasting. However, LLMs have recently emerged in the arena of time series forecasting and have made significant progress in various fields such as healthcare, finance, and transportation \cite{LLM_in_TS}. Time-LLM \cite{Time_LLM} proposed a novel framework repurposing LLMs for time series forecasting without requiring any fine-tuning of the backbone model. This was achieved by \textquote{reprogramming} time series data inputs for compatibility with LLMs; thereby, converting time series forecasting into a \textquote{language} task. An LLM's advanced reasoning and pattern recognition capabilities could then be leveraged to achieve high precision and efficiency in forecasts. Time-LLM was shown to outperform specialized models in few-shot and zero-shot scenarios. 

Similarly, Chronos \cite{Ansari2024ChronosLT} proposed the use of LLMs for time series forecasting. However, it avoided reprogramming the time series data, which requires training on each input dataset separately. Instead, time-series data was tokenized into a fixed vocabulary via scaling and quantization. The Chronos model outperformed statistical baselines and other pretrained models in both in-domain and zero-shot scenarios across multiple benchmarks. 

LLMTime \cite{LLM_zero_shot_TS} also proposed the use of LLMs for time series forecasting. Rather than requiring learned input transformations or prompt engineering as Time-LLM did, time series data were tokenized like with Chronos but with a different scheme. In fact, for this framework, effective numerical tokenization was essential in ensuring accurate and efficient forecasting by the LLMs. LLMTime outperformed traditional statistical models and models from the Monash forecasting archive. Furthermore, it was competitive with and sometimes outperformed efficient transformer models. 

PromptCast \cite{Xue2022PromptCastAN} also introduced a novel approach to time series forecasting using LLMs. Like Time-LLM, numerical sequences are described and transformed to natural language sentences. However, PrompCast used manually-defined template-based prompting rather than learning input transformations for automatic prompting. While explored for only unistep forecasting, the results indicated that the PromptCast approach not only achieved performance that was comparable to traditional numerical methods but sometimes even surpassed them. 

These prior works suggest the emergence of multimodal models that excel in both language and time series forecasting tasks. However, these works presented LLMs for use in only time series forecasting and did not explore other time series tasks like anomaly detection. However, in separate works, LLMs have emerged for other time series tasks and have been shown to excel. Time series tasks typically include four main analytical tasks: forecasting, classification, anomaly detection, and imputation \cite{LLM_in_TS}. 

Zhou et al. \cite{Zhou2023OneFA} introduced a unified framework (referred to as One Size Fits All (OFA) \cite{LLM_in_TS}) that uses frozen pretrained LLMs for performing various time series analysis tasks. Like Time-LLM, OFA required training the input embedding layer to acquire learned time series representations. However, rather than only time series forecasting, it explored the use of LLMs for univariate anomaly detection. OFA achieved superior or comparable results in classification, forecasting, anomaly detection, and few-shot/zero-shot learning. 

Sun et al. \cite{Sun2023TESTTP} proposed an embedding method for TimE Series tokens to align the Text embedding space of LLM (TEST). TEST's embeddings alignment methodology enhances LLMs' ability to perform time series tasks without losing language processing abilities. Although the exact embedding function was not specified, learning input transformations typically involves neural network training. Therefore, like Time-LLM, TEST also required training the input embedding layer. However, like OFA, TEST explored the use of LLMs for other time series tasks. Compared to state-of-the-art models, TEST demonstrated superior performance on various tasks including univariate time series forecasting, as well as multivariate classification tasks. 

While achieving good performance on multiple time series tasks, neither OFA nor TEST explored multivariate anomaly detection. Multivariate analysis allows for joint reasoning across the time series. Joint reasoning enables a model to blend and merge the understanding from different sensors and data sources to make decisions that are impossible when considering data in isolation. For example, in our use-case, the temperature alone may not sufficiently indicate a problem since operators might adjust the temperature to try and maintain material flow despite a screen pack blockage. By monitoring both pressure and temperature, it is possible to detect joint anomaly events that are more indicative of clogging. Furthermore, there were no papers exploring LLMs for the PdM use-case.

\begin{figure*}
    \includegraphics[width=0.95\textwidth, height=3.5in]{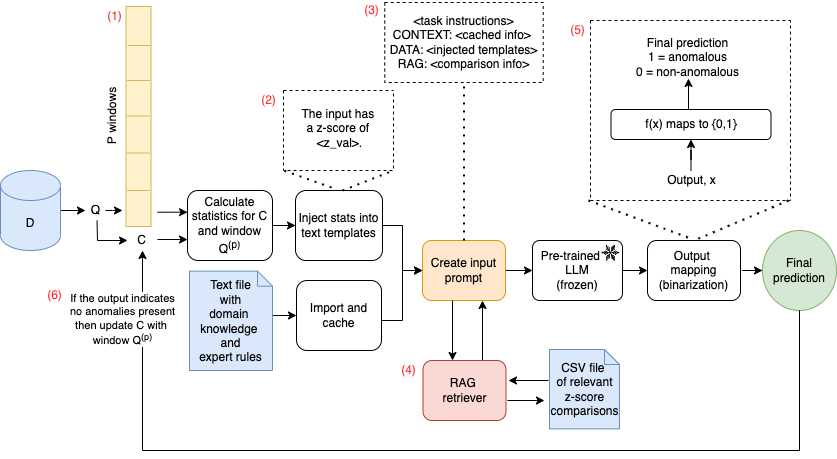}
    \caption{The model framework of RAAD-LLM. Given an input time series $Q$ from the dataset $D$ under consideration, we first preprocess it using SPC techniques. Then \textcolor{red}{(1)} $Q$ is partitioned into a comparison dataset $C$ and query windows $Q^{(p)}$, where $p \in P$ and $P$ is the number of segmented windows. Next, statistical measures for $C$ and $Q^{(p)}$ are calculated and \textcolor{red}{(2)} injected into text templates. These templates are combined with task instructions to create the input prompt. To enhance the LLM's reasoning ability, \textcolor{red}{(3)} domain context is added to the prompt. Statistical measures for all input variables are sent to the RAG component \textcolor{red}{(4)} to retrieve relevant z-score comparison information from the knowledge base. Retrieved information is combined with the prompt before being fed forward through the frozen LLM. The output from the LLM is \textcolor{red}{(5)} mapped to $\{0,1\}$ via a binarization function to obtain the final prediction. \textcolor{red}{(6)} Updates to $C$ are determined before moving to the next $Q^{(p)}$.}
    \label{fig:fig_meth1}
\end{figure*}

\subsection{Enriching Time-Series Data With Semantics for AAD in PdM}
Advancement in anomaly detection through adaptability has been explored extensively. Traditionally, most AAD algorithms have
been designed for data sets in which all observations are available at one time (i.e., static datasets). However, over the last two decades, many algorithms have been proposed to detect anomalies in "evolving" data (i.e., data streams) \cite{Salehi2018ASO}. Although the proposed methodology could possibly be modified for data streams, we only focus on static datasets in this paper. 

ML and NN techniques have been used for AAD implementation and have been shown to improve the performance baselines of non-adaptive models in various scenarios such as industrial applications \cite{Singh2023AdaptiveAD}, network security \cite{Xu2023ADTCDAA}, and environmental science \cite{Salehi2018ASO}. However, these techniques focus only on the data themselves. Although effective, these approaches may overlook contextual information and domain-specific knowledge crucial for accurate anomaly detection. 

A system that combines ML and semantics improves the accuracy of anomaly detection in the data by reducing the number of false positives \cite{Steenwinckel2018AdaptiveAD, Steenwinckel2021FLAGSAM}. This is because integrating semantics into the anomaly detection process allows for a more comprehensive analysis that considers both the data patterns and their contextual relevance. A system like this would enable for more collaborative decision-making between the model and plant operators.

Semantics such as the following could greatly enhance anomaly detection, as it provides insight into the severity of the anomaly: 
\begin{displayquote}
Domain-specific knowledge indicates that there are correlations between process variables. Specifically, increased melt pressure at the screen pack inlet may lead to increased melt temperature at the screen pack inlet. Additionally, increased melt pressure at the screen pack inlet may lead to decreased melt pressure at the screen pack outlet. If these correlations are observed, it indicates a high level of criticality for potential failures.
\end{displayquote}
In this case, plant operators may want to imply that if an anomaly is not severe enough, then it is a false positive; and therefore, should not trigger a manual shutdown. Unlike ML models, LLMs can easily integrate this knowledge for the anomaly detection task.

Few previous works have incorporated expert knowledge with ML algorithms for anomaly detection within time-series data. Ontology-based LSTM (OntoLSTM) \cite{HUANG2019437} integrates ontology-driven representations with DL to model manufacturing time-series data. Its framework combines a hierarchical ontology-based NN with stacked dense layers for \textquote{learning} representations of manufacturing lines and machines, and an LSTM module for capturing temporal dependencies in production process data. Adaptability in OntoLSTM stems from its ability to dynamically integrate domain-specific semantics into its deep architecture, allowing it to align with varying manufacturing processes. However, the model requires extensive training due to its hybrid nature, as it must optimize both the representation-learning dense layers and the LSTM’s temporal learning component to accurately detect anomalies.

Fused-AI interpretabLe Anomaly Generation System (FLAGS) \cite{Steenwinckel2021FLAGSAM} integrated data-driven and knowledge-driven approaches to deliver adaptive, context-aware anomaly detection. The Semantic Mapping module is responsible for enriching the incoming data streams with expert rules and context information. Adaptability here refers to the merging, deleting, or relabeling of anomalies to cope with user-provided feedback; and dynamic rule extracting. FLAGS is an ensemble architecture that uses one ML model to detect anomalies and another that fuses semantics to determine whether they are true anomalies. Although the FLAGS architecture allows for the use of any appropriate ML models, non-LLM models are largely statistical without much innate reasoning \cite{Time_LLM}. 

Notably, LLMs demonstrate advanced abilities in reasoning and data synthesis \cite{wang2024enhancing, chu2023leveraging}, and offer few/zero shot capabilities and transferability \cite{LLM_few_shot, LLM_few_shot_health, LLM_few_shot_prog}. Since pretrained LLMs have been shown to perform well on various time-series tasks, leveraging their learned higher level concepts could enable highly precise and synergistic detection across multiple modalities \cite{Time_LLM}.  Furthermore, while traditional ML or NN models typically require more specialized training, LLMs have the ability to perform well with less data and without extensive retraining. This is extremely advantageous in data-constrained operational settings.

%% file: component_subject_sections/Methodology/methodology.tex
This section overviews the RAAD-LLM methodology and outlines the proposed enhancement of the AAD-LLM framework through the incorporation of a RAG pipeline. The goal of this enhancement is to improve the model’s capacity to perform complex reasoning tasks that require computational support. The integration of the RAG pipeline into the AAD-LLM framework allows the model to access external knowledge bases dynamically and incorporate relevant information into its decision-making process. This combination enhances the LLM's performance in recognizing and classifying anomalies. As a result, this integration will strengthen RAAD-LLM's ability to detect anomalies in data-sparse industrial environments.

\subsection{The RAAD-LLM Framework}

The following subsections provide a detailed description of the RAAD-LLM architecture as shown in \autoref{fig:fig_meth1}. It discusses the key components, domain-specific knowledge integration, data processing workflow, and the methodology for combining prompting, RAG, and model inference to enhance anomaly detection capabilities.

\subsubsection{System Architecture}

The integrated architecture, referred to as RAAD-LLM, consists of the following key components:
\begin{enumerate}
\item \textit{Pretrained LLM:} A pretrained LLM serves as the foundation model for anomaly detection. For this work, Meta Llama 3.1 8B model was chosen since it is an open source instruction-fine-tuned LLM demonstrating state-of-the-art on tasks such as classification, question answering, extraction, reasoning, and summarizing \cite{Zhang2023InstructionTF}. This model remains frozen during processing to maintain transferability.
\item \textit{RAG Component:} The RAG component incorporates a CSV-based knowledge base that dynamically retrieves contextually relevant information for z-score comparisons. This functionality enables the model to provide responses reflecting either exact matches for input values or the closest available matches when exact values are not found. Consequently, the integration of RAG improves both the accuracy and interpretability of sensor data queries.
\item \textit{Anomaly Detection Layer:} For each time series window under consideration, this layer applies SPC techniques and the Discrete Fourier Transform (DFT), and then computes historical and current statistical measures. Next, utilizing domain-specific knowledge, text templates, and a binarization function anomaly classification is performed.
\item \textit{Adaptability Mechanism:} This feature continuously updates the baseline normal behavior as each new time series window is processed. 
\end{enumerate}

The LLM hosting environment is built around Ollama, which supports the Llama 3.1 8b model. This server acts as the central endpoint for processing context-aware prompts. The server’s base URL is specified, and parameters like request timeout and maximum tokens are configured to ensure steady communication and manage resource usage.

\subsubsection{Domain-Specific Knowledge and Text Templates}

To enhance collaboration with plant operators, we develop a domain-specific context file that enables the LLM to comprehend the specifics of our time series data. This file integrates expert rules, domain knowledge, and constraints to establish acceptable ranges for process variable variations, guide feature selection, and describe in detail causal relationships among variables. In our manufacturing use case, comprising 580 sensors per line, operators can correlate these readings with failure modes. Furthermore, fluctuations in raw materials necessitate adjustments in process parameters, which polymer scientists can specify. By utilizing this expertise, we can refine thresholds, select relevant features, and identify interactions; thereby improving anomaly detection. The context file is imported and should be persisted for efficiency. 

To enable structured understanding and improved performance, we create text templates with placeholders that align with essential statistical values such as mean, standard deviation, and maximum. In this work, only the z-score is used, as prior research found it sufficient to yield good results. When actual data is available, these placeholders are populated through data injection. Injected statistical measures for both normal system behavior and the current query window under consideration will guide the LLM’s reasoning, enhancing its anomaly detection capabilities.

\begin{center}
    \begin{minipage}{9cm}            
        \begin{tcolorbox}[enhanced,attach boxed title to top center={yshift=-1mm,yshifttext=-1mm},
                colback=green!10!white,colframe=gray!90!black,colbacktitle=gray!80!black, left=0.1mm, right=0.5mm, boxrule=0.50pt]
        \textbf{INSTRUCTIONS:} You are a helpful assistant that can use these rules to answer queries. The following sensor data was collected over the last 15 minutes and represent current process conditions. Strictly based on the context and RAG information provided below, please answer the following questions. Do not modify, interpret, or apply logic beyond these instructions. \\ 
        * Is high deviation present for Melt Pressure 1? \\
        * Is high deviation present for Melt Pressure Differential? \\
        For each question, avoid explaining. Just print only the output and nothing else. \\        
        \textbf{CONTEXT:} \textcolor{red}{$<$cached info$>$} \\
        \textbf{DATA:} Melt Pressure 1 has a z-score of \textcolor{red}{$<$val$>$}. Melt Pressure Differential has a z-score of \textcolor{red}{$<$val$>$}. \\
        \textbf{RAG:} The z-score for Melt Pressure 1 is \textcolor{red}{$<$greater than / less than / equal to$>$} acceptable process variable conditions. The z-score for Melt Pressure Differential is \textcolor{red}{$<$greater than / less than / equal to$>$} acceptable process variable conditions.

        \end{tcolorbox}
    \end{minipage}
\end{center}
\vspace{0mm}
\captionof{figure}{Prompt example. \textcolor{red}{$<$cached info$>$} is the domain context information. \textcolor{red}{$<$val$>$} are calculated statistical measures injected into respective text templates. \textcolor{red}{$<$greater than / less than / equal to$>$} is the relevant z-score comparison information from the RAG retriever. Note that although each $Q_i$ is processed independently, prompts include text templates for all $i \in N$ where $N$ is the number of input variables in instance $Q$ from the dataset $D$ under consideration. Therefore, multivariate anomaly detection is explored. }\label{fig:prompt_text}
\vspace{4mm}

\setlength\parindent{12pt}
\subsubsection{Data Processing Workflow}

Defining normal process behavior is crucial for effective anomaly detection, as it establishes a baseline against which potential anomalies can be compared and identified. This baseline is determined in a manner akin to the AAD-LLM methodology. From the dataset D under consideration, a multivariate time series instance $Q\in \mathbb{R}^{N\times T}$ is partitioned into $N$ univariate time series where $N$ is the number of input variables and $T$ is the number of time steps. This is done so that each input variable is processed independently \cite{Time_LLM}. 

Each $i^{th}$ series $Q_i, i \in N$, is then processed using SPC techniques. For this work, the univariate MAMR charts are plotted for each process variable as shown in \autoref{fig:fig_meth3}. This aspect of plotting and analyzing the MAMR charts for all process variables in parallel will cause an increase in the Type I error rate. Time series points deemed \textquote{out of statistical control} are labeled as anomalous and filtered out of $Q_i$ before further processing. SPC is applied again after the first set of outliers (or anomalies) are removed. This is done to ensure extreme values do not affect control limits. Therefore, it can be assumed that time series $Q_i$ represents a stable process. We use this assumption in initializing our comparison dataset $C_i$ as our baseline for normal behavior as explained in the next paragraph. The idea is that once the comparison dataset is initialized, the model then updates its understanding of normalcy as each new query window is ingested.

Rather than processing the entire time series at once, $Q_i$ then undergoes windowing as shown in \autoref{fig:fig_meth1}. For each $i \in N$, windowing divides time series $Q_i$ into $P$ consecutive non-overlapping segments of length $L$, $Q_i^{(P)} \in \mathbb{R}^{P \times L}$. By analyzing data within sliding windows, anomaly detection can focus on smaller segments of the time series data. This provides a more granular and detailed view of abnormal patterns. Processing the entire time series as a single entity might obscure localized anomalies within the data. Finally, for each $i \in N$, a baseline dataset $C_i \in \mathbb{R}^{1 \times L}$ of normal behavior is defined as the first $Q_i$ window.  

Unlike AAD-LLM, for each $i \in N$, both the current $Q_i$ window $Q_i^{(p)}$ where $p \in P$ and the baseline data set, $C_{i}$ undergo additional processing through the DFT. Since sensor signals are sampled discretely, the DFT is very useful in their analyses \cite{e2019troubleshooting}. Specifically, the DFT can be used to isolate the most prominent frequency component from the noise, thereby enhancing the discernibility of the signal. After applying the DFT, we then construct a sinusoidal representation of the dominant frequency component for both $ Q_i^{(p)} $ and $ C_i $. The steps to apply the DFT and then construct a sinusoidal representation are as follows.

The DFT for signal $s(t)$ is computed as:
\begin{equation}
    F(k) = \sum_{t=0}^{N-1} s(t) \cdot e^{-i \cdot 2\pi \frac{k \cdot t}{N}} \quad \text{for } k = 0, 1, \dots, \frac{N}{2}
\end{equation}
Here, $F(k)$ represents the frequency components, with the focus on the real part of the spectrum.

The amplitude spectrum is computed as the scaled magnitude of the Fourier coefficients:
\begin{equation}
A_k = \frac{2}{N} \left| F_k \right| \quad \text{for } k = 0, 1, \dots, \frac{N}{2}
\end{equation}
where:
\begin{itemize}
    \item $A_k$ is the amplitude corresponding to the $k$-th frequency component,
    \item $F_k$ is the $k$-th Fourier coefficient from the DFT output.
\end{itemize}

The dominant frequency and amplitude are determined as:
\begin{equation}
f_{\text{max}} = f_k \quad \text{where } k = \arg\max_k A_k
\end{equation}

\begin{equation}
A_{\text{max}} = A_k \quad \text{for the same } k
\end{equation}
where:
\begin{itemize}
    \item $f_{\text{max}}$ is the dominant frequency in the signal,
    \item $A_{\text{max}}$ is the amplitude of the dominant frequency.
\end{itemize}

Using $f_{\text{max}}$ and $A_{\text{max}}$, a sine wave is fitted to represent the dominant signal component:
$$
\hat{s}(t) = A_{\text{max}} \cdot \sin(2 \pi f_{\text{max}} \cdot t) + \left| \bar{s} \right|
$$
where:
\begin{itemize}
    \item $\hat{s}(t)$ is the reconstructed sine wave signal,
    \item $A_{\text{max}}$ is the amplitude of the dominant frequency,
    \item $f_{\text{max}}$ is the dominant frequency,
    \item $\bar{s}$ is the mean value of the original signal that is added to account for offset adjustments,
    \item $t$ represents time.
\end{itemize}

Subsequently, selected statistical measures for the sinusoidal representations of $Q_i^{(p)}$ and $C_i$ are calculated and then injected into the corresponding text templates. This approach is advantageous because it allows for a clearer differentiation between signal and noise, making it easier to identify patterns and anomalies in the data. By focusing on frequency components, we gain a deeper understanding of the underlying dynamics of the signal.

\subsubsection{Prompting, RAG, and Model Inference} \label{sec:prompting}

Prompts are then created via prompt engineering and combined with the templates. To further enrich the inputs, the domain context is added to the prompt before being fed forward through the frozen LLM. For our methodology, the domain context was manually restructured from the \textquote{raw} domain context to reduce the complexity of the input prompt. Consequently, this better guided the LLM's decision making, thereby enabling more consistent predictions. Effective prompt engineering is essential in ensuring accurate, context-aware anomaly detection.

Prior to predicting anomalies, the statistical measures for all input variables are sent to the RAG component to retrieve relevant z-score comparison information from the knowledge base. The retrieved information is then combined with the prompt, allowing the LLM to better understand the relationship between the historical normal and the observed statistics of the process being monitored. A prompt example is shown in \autoref{fig:prompt_text}. The resultant enriched prompt is fed forward through the frozen LLM. 

\begin{center}
    \begin{minipage}{9cm}            
        \begin{tcolorbox}[enhanced,attach boxed title to top center={yshift=-1mm,yshifttext=-1mm},
                colback=green!10!white,colframe=gray!90!black,colbacktitle=gray!80!black, left=0.1mm, right=0.5mm, boxrule=0.50pt]
        * High deviation is present for Melt Pressure 1. \\
        * High deviation is not present for Melt Pressure Differential.    

        \end{tcolorbox}
    \end{minipage}
\end{center}
\vspace{0mm}
\captionof{figure}{LLM output example. Outputs are an itemized list of process variables and their anomaly status. The text-based outputs use domain-specific terminology, enabling subject matter experts to interpret findings more easily than numerical results and fostering better collaboration and knowledge transfer.}\label{fig:llm_output}
\vspace{4mm}

\setlength\parindent{12pt}
An example output of the LLM is shown in \autoref{fig:llm_output}. The LLM outputs an itemized list indicating whether an anomaly is present for each process variable. The textual outputs of the LLM enhance collaboration with subject matter experts because they are more accessible and easier to interpret than purely numerical results. These text-based outputs incorporate domain-specific terminology that allow experts to understand findings without the need to decode complex numbers. This enhancement fosters better communication and feedback loops between technical and non-technical team members. Consequently, experts can validate or challenge the model’s conclusions more effectively. Ultimately, this approach promotes improved knowledge transfer and bridges the gap between expert systems and domain expertise, making the outputs significantly more actionable and user-friendly.

Lastly, we apply a binarization function to the LLM's outputs to map them to $\{0,1\}$ to get the final classification ($0=$ non-anomalous, $1=$ anomalous). The exact binarization function is use-case specific. For our use-case, one anomaly alone does not sufficiently indicate a problem. To avoid false positives that trigger an unnecessary shutdown, our binarization function only maps to $1$ if anomalies in the output are correlated as indicated by domain-specific knowledge. Let $x$ be the LLM output. Then

\begin{equation}
     f(x) =
    \begin{cases} 
        1, & \text{if anomalies in } x \text{ are correlated} \\
        0, & \text{otherwise}
     \end{cases}
\end{equation}

The final classification is what is used for determining updates to $C_i$ before moving to the next $Q_i^{\left(p\right)}$. If the output prediction indicates no anomalies in $Q_i^{\left(p\right)}$, window $Q_i^{\left(p\right)}$ series data is combined with the preceding windows series data to gradually refine the dataset of what constitutes \textquote{normal} behavior $C_i$. Therefore, for each $i \in N$, $C_i$ is guaranteed to be representative of normal behavior and is constantly evolving.

\subsubsection{Adaptability Mechanism}

The adaptability mechanism of AAD-LLM is preserved in the RAAD-LLM framework. In addition to $C_i$ constantly updating as each new query window is ingested, the process of re-initializing $C_i$ is done for each new instance $Q$. This continuous redefining of the normal baseline enables the model to progressively refine its knowledge in response to shifts in the system’s operational conditions process after process. Therefore, the model is enabled to maintain an up-to-date and broad perspective of normality. 

\begin{table*}[!t]
    \begin{center}
    \begin{tabular}{| c | c |}
    \hline
    \textbf{Component} & \textbf{Details} \\
    \hline
    \hline
    LLM Model &  Meta Llama 3.1 model (8B parameters) \\
    \hline
    Server Base URL & Hosted on a private network at http://localhost:11434 \\
    \hline
    Request Timeout & 500 seconds \\
    \hline
    Output Temperature & 0.2 \\
    \hline
    Token Limit & 250 tokens \\
    \hline
    Mirostat & Disabled (mirostat: 0), ensuring deterministic output generation \\
    \hline
    Embedding Use & The LLM is utilized to generate embeddings for domain-specific vector stores for efficient context retrieval. \\
    \hline
    \end{tabular}
    \caption{Configuration summary table of the LlamaIndex and Ollama system. This setup facilitates seamless retrieval of relevant domain knowledge from the vector store using LlamaIndex. Rather than all the domain context, only the retrieved content is added to the prompt before being fed forward through the frozen LLM.} 
    \label{table:config}
    \end{center}
\end{table*}

\begin{figure}
    \includegraphics[width=0.48\textwidth]{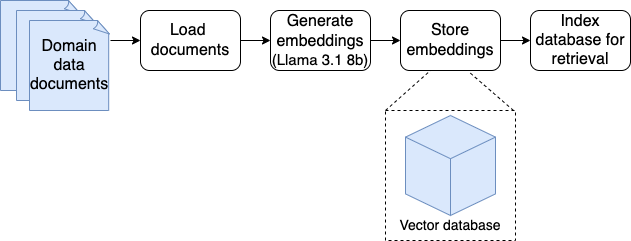}
    \caption{The LlamaIndex flowchart representation.  Raw domain context information is loaded as input. Each data chunk is processed using an embedding model (in this case, LLama 3.1 8b from the Ollama server). Parameters such as temperature (0.2), max tokens (250), and mirostat (disabled) are set to ensure robust and consistent embeddings are generated for the context. The generated embeddings are then stored as vectors in a vector database. Finally, LlamaIndex organizes and indexes the embeddings into a retrievable format. The vector store then becomes accessible to the RAG component, allowing dynamic retrieval of relevant context as needed.}
    \label{fig:fig_llamaindex}
\end{figure}

\subsection{The RAAD-LLMv2 Framework With LlamaIndex Integration}

The RAAD-LLM architecture requires that all domain context be added to the prompt before being fed through the frozen LLM. This made the query too complex, leading to inconsistent responses that often did not align with expectations. To address this issue, we manually restructured the \textquote{raw} domain context as described in \autoref{sec:prompting}. This restructuring better guided the LLM's decision-making, but it took a lot of time and effort.

The RAAD-LLMv2 variant extends RAAD-LLM by integrating an additional RAG module powered by LlamaIndex.  LlamaIndex is open-sourced and has been proposed as a method to expand the context capabilities of LLMs by enabling them to utilize extensive documents and databases during response generation \cite{zirnstein2023extended, malviya2024scalability, malviyanew}. The new architecture dynamically retrieves relevant domain context rather than incorporating all the provided context into the prompt. Consequently, this new architecture enhances the model's decision-making by providing more accurate and consistent responses without manual context restructuring. Additionally, it is more scalable under real-world scenarios. \autoref{fig:fig_llamaindex} is a visual representation of the LlamaIndex process.

\subsubsection{LlamaIndex and Ollama System Configuration}

The configuration of the LlamaIndex and Ollama system is designed to enable effective interaction between the RAG component and the LLM for domain context retrieval and embedding generation. The following provides technical details about the integration, including the configuration of the LlamaIndex and Ollama system. 

\begin{enumerate}
\item \textit{Ollama LLM Server:} The LLM hosting environment is the same as for the RAAD-LLM framework. This server acts as the central endpoint for processing both context-aware prompts and embeddings for vector stores. 
\item \textit{Embedding Model:} The LlamaIndex relies on Ollama's embedding capabilities, using the same Llama 3.1 model as an embedding generator for the knowledge base. 
\item \textit{Parameter Tuning:} Both the LLM and embedding configurations include custom parameters optimized to balance accuracy and computational efficiency. These parameters govern model output behaviors, such as temperature (for controlling randomness), maximum token count (to limit the size of outputs), and the request timeout duration.
\end{enumerate}

\autoref{table:config} details the LlamaIndex and Ollama system configuration used for this work. This configuration facilitates the retrieval of relevant information from a vector store to complement input prompts, thereby improving the LLM's contextual understanding.

%% file: component_subject_sections/Results/results.tex
This section discusses the analyses of the datasets and experimental outcomes of RAAD-LLM and RAAD-LLMv2. The focus is on their performance improvements and limitations when applied to anomaly detection tasks. These discussions aim to provide deeper insights into the frameworks' effectiveness and areas for future enhancement.

\subsection{Data and Analysis for the PdM Use-Case}

\begin{figure}
    \includegraphics[width=0.45\textwidth]{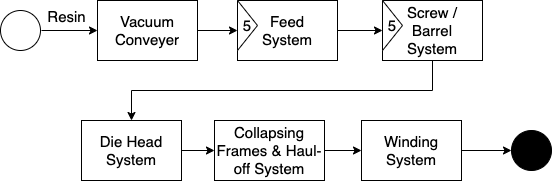}
    \caption{Process flow diagram of major components in our use-case extrusion process. The major components in the extrusion process are in a series configuration. The number of Feed and Screw/Barrel Systems depends on the manufacturing line number and can be 3, 4, or 5.}
    \label{fig:fig_process}
\end{figure}

\begin{figure}
    \includegraphics[width=0.5\textwidth]{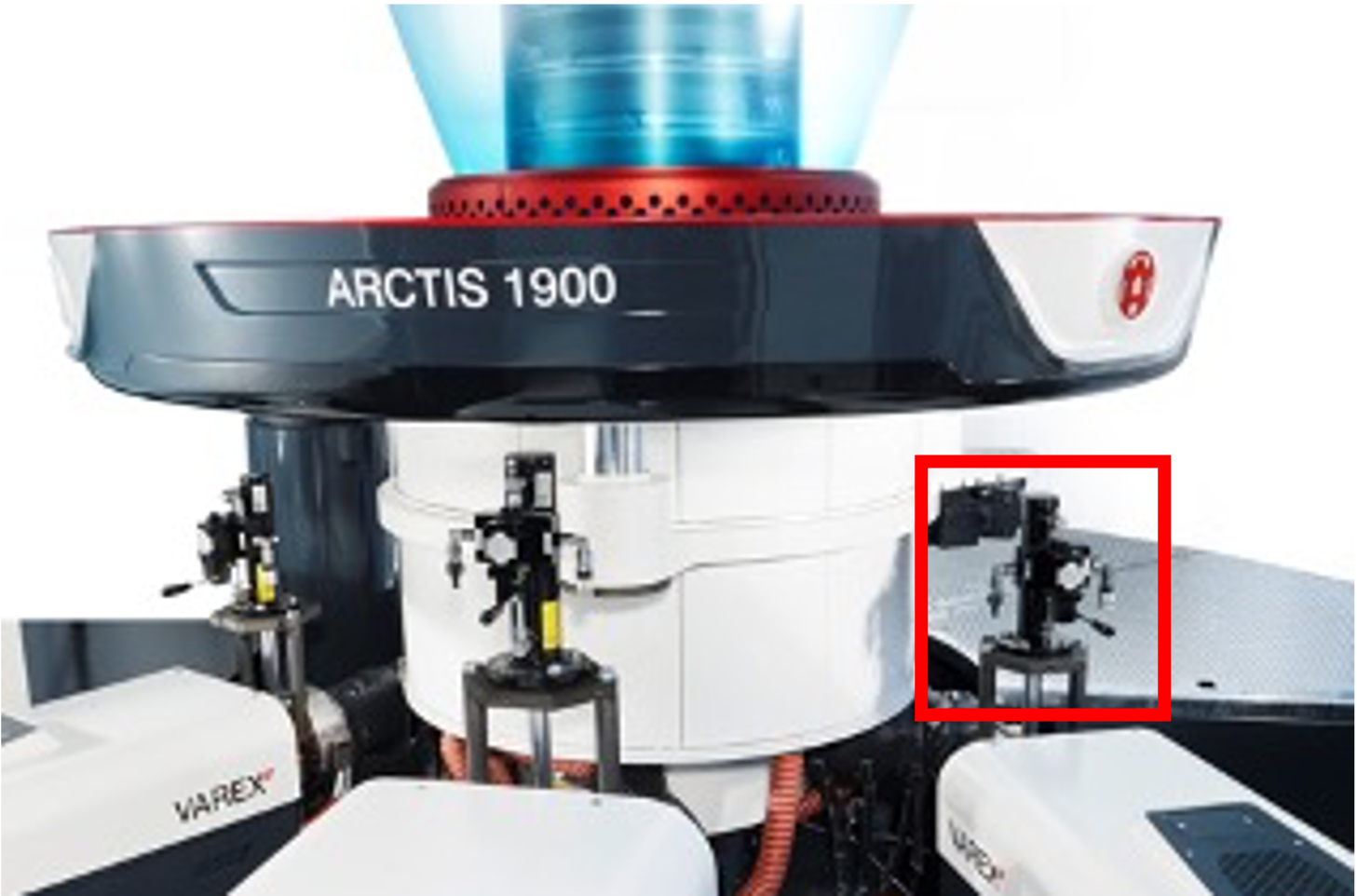}
    \caption{The die head system for our use-case. The screen pack changer is identified by a red box. Within the screen pack changer, screens are used to prevent impurities from getting into the extruder together with the resin and thus clogging the die gap. The number of screen packs depend on the number of Screw/Barrel Systems. Each screen pack is arranged between the Screw/Barrel System and the Die Head System. During production, the resin melts flow through the screen pack.}
    \label{fig:fig_screenpack}
\end{figure}

Our use-case dataset was for screen pack failures in the extrusion process since shutdowns due to these failures were well documented by the plastics manufacturing plant providing the data. An example of a screen pack changer can be seen in \autoref{fig:fig_screenpack} and an overview of the plastics extrusion process for our use-case can be seen in \autoref{fig:fig_process}. For two downtime events with screen pack failure mode, we obtained 65 hours of historical run-to-failure sensor readings (6.5 hours for 5 components for each downtime event). The readings were semi-labeled and for process variables that were deemed good indicators of screen pack failures. These process variables are \textit{Melt Pressure 1}, \textit{Temperature 1}, and \textit{Melt Pressure Differential}. 

\begin{itemize}
\item Melt Pressure 1 - The melt viscosity at the screen inlet. 
\item Temperature 1 - The melt temperature at the screen pack inlet. 
\item Melt Pressure Differential - The melt pressure across the screen pack inlet and outlet. 
\end{itemize}
For any of these, sudden spikes from expected profile could signal significant process variable deviations; and therefore, could lead to a screen pack failure. Since \textit{Temperature 1} did not contain enough sample data, it was not used for input into RAAD-LLM and RAAD-LLMv2 for anomaly detection.

The domain context was meticulously collected from maintenance logs and plant operators. Maintenance logs provided detailed records of prior screen pack failures and anomalies. Additionally, plant operators contributed their expertise and firsthand knowledge, which helped define acceptable ranges for fluctuations and establish causal relationships among process variables. This collaborative approach ensured that the domain context effectively captured the operational intricacies of the manufacturing process.

\begin{figure}
    \includegraphics[width=0.5\textwidth]{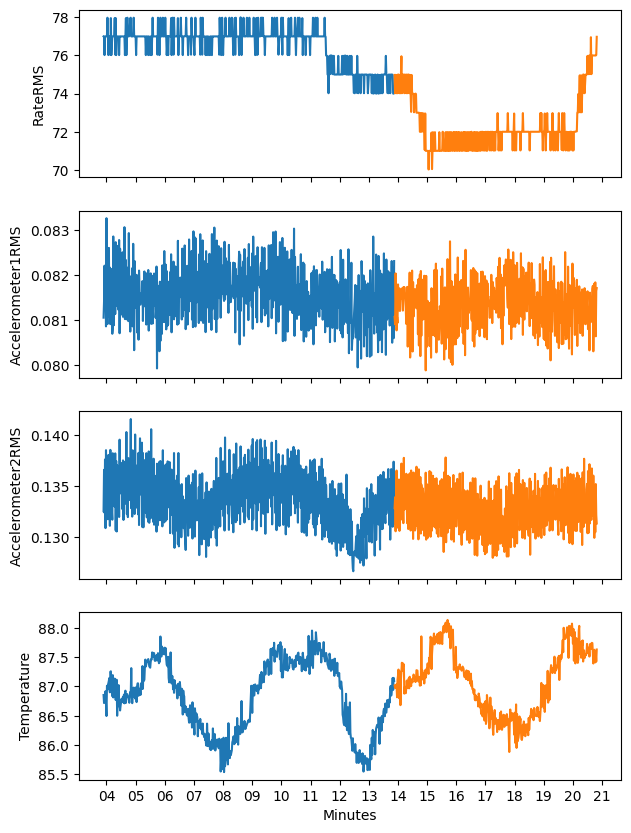}
    \caption{Processed sensor data from the SKAB dataset. The selected signals are preprocessed to include only those experiments that were 20 minutes in duration. The first 3 minutes were discarded as process start-up. Each signal begins in a non-anomalous experimental state and continues until the end of the experiment. Non-anomalous states are shown in \textcolor{blue}{blue} and anomalous states are shown in \textcolor{orange}{orange}. Processed signals are then input into the frameworks.}
    \label{fig:fig_sensors}
\end{figure}

\subsection{Data and Analysis for the SKAB Dataset}

The Skoltech Anomaly Benchmark (SKAB) is a publically accessible dataset designed for evaluating the performance of anomaly detection algorithms. The benchmark includes labeled signals captured by several sensors installed on the SKAB testbed. The SKAB testbed was specifically developed to study anomalies in a testbed. The focus of this work is to develop methods for detecting anomalies in these signals, which can be relevant for various applications. 

A description of the columns in the SKAB dataset is as follows \cite{skab}.
\begin{itemize}
    \item datetime - Dates and times when the value collected
    \item Accelerometer1RMS - Vibration acceleration (g units)
    \item Accelerometer2RMS - Vibration acceleration (g units)
    \item Current - The amperage on the electric motor (Ampere)
    \item Pressure - The pressure in the loop after the water pump (Bar)
    \item Temperature - The temperature of the engine body ($\degree C$)
    \item Thermocouple - The temperature of the fluid in the circulation loop ($\degree C$)
    \item Voltage - The voltage on the electric motor (Volt)
    \item RateRMS - The circulation flow rate of the fluid inside the loop (Liter per minute)
    \item anomaly - If the point is anomalous (0 or 1)
    \item changepoint - If the point is a changepoint (0 or 1)
\end{itemize}

The anomaly column contains the labels. A Mann–Whitney–Wilcoxon test was used to determine whether any of the data features affected the labels. This test combined with a correlation matrix to detect relationships between variables resulted in \textit{Accelerometer1RMS}, \textit{Accelerometer2RMS}, \textit{Temperature}, and \textit{RateRMS} as selected inputs for RAAD-LLM and RAAD-LLMv2 to make the predictions. See \autoref{fig:fig_sensors} for further processing details.

Experiments on the SKAB dataset were conducted to determine the optimal fluctuation ranges for each of the selected features. Domain context was determined without using any prior domain knowledge. As a result, the context was built solely on a single statistical measure, which may have limited the accuracy of anomaly detection in this highly specialized system. Incorporating domain expertise could have enabled better feature selection, threshold setting, and understanding of variable interactions. Consequently, the model’s performance on the SKAB dataset may have been constrained, highlighting the potential for improvement through informed context creation. See \autoref{sec: perf} for model results on SKAB.

\subsection{Evaluation of Model Performance} \label{sec: perf}

\begin{table}[!h]
    \begin{center}
    Use-Case Dataset \\
    \begin{tabular}{| c | c | c | c | c |}
        \hline
        \textbf{Model} & \textbf{Accuracy} & \textbf{Precision} & \textbf{Recall} & \textbf{F1 score}\\
        \hline
        \hline
        Baseline & 0.74 & 0.74 & 1.00 & 0.85 \\
        \hline
        AAD-LLM & 0.71 & 0.88 & 0.68 & 0.77 \\
        \hline
        RAAD-LLM & 0.89 & 0.93 & 0.91 & 0.92 \\
        \hline
        RAAD-LLMv2 & 0.73 & 0.96 & 0.66 & 0.78 \\
        \hline
        \end{tabular}
        \bigskip
        
        SKAB Dataset \\
        \begin{tabular}{| c | c | c | c | c |}
        \hline
        \textbf{Model} & \textbf{Accuracy} & \textbf{Precision} & \textbf{Recall} & \textbf{F1 score}\\
        \hline
        \hline
        Baseline & 0.45 & 0.45 & 1.00 & 0.62 \\
        \hline
        AAD-LLM & 0.58 & 0.47 & 0.68 & 0.56 \\
        \hline
        RAAD-LLM & 0.72 & 0.63 & 0.89 & 0.74 \\
        \hline
        RAAD-LLMv2 & 0.68 & 0.61 & 0.81 & 0.70 \\
        \hline
    \end{tabular}
    \vspace{4mm}
    \caption{Average evaluation metrics over the best 5 model runs. The baseline model is one that predicts every observation to belong to the positive class. The RAG pipeline for both RAAD-LLM and RAAD-LLMv2 integrate a CSV-based knowledge base to dynamically retrieve relevant information for z-score comparisons, allowing for responses that reflect either exact matches for input values or the closest matches when exact values are not found. RAAD-LLMv2 integrates LlamaIndex for seamless retrieval of relevant domain knowledge from the vector store. Unlike RAAD-LLM, RAAD-LLMv2 adds only the retrieved content to the prompt before being fed forward through the frozen LLM.} \label{table:tab_results1}
    \end{center}
\end{table}

\begin{table*}[!t]
    \begin{center}
    \begin{tabular}{| c | c | c | c | c | c | c | c |}
    \hline
    \textbf{Algorithm} & \textbf{F1} & \textbf{FAR, \%} & \textbf{MAR, \%} & \textbf{No Training or Fine-tuning} & \textbf{Multimodal} \\
    \hline
    \hline
    Perfect detector & 1 & 0 & 0 & & \\
    \hline
    \textbf{RAAD-LLM} & \textbf{0.74} & \textbf{42.05} & \textbf{11.43} & \textbf{yes} & \textbf{yes} \\
    \hline
    LSTMCaps \cite{Elhalwagy_2022} & 0.74 & 21.5 & 18.74 & no & no \\
    \hline
    MSET \cite{mset} & 0.73 & 20.82 & 20.08 & no & no \\
    \hline
    LSTMCapsV2 \cite{Elhalwagy_2022} & 0.71 & 14.51 & 30.59 & no & no \\
    \hline
    \textbf{RAAD-LLMv2} & \textbf{0.70} & \textbf{42.05} & \textbf{18.67} & \textbf{yes} & \textbf{yes} \\
    \hline
    MSCRED \cite{Zhang_Song_Chen_Feng_Lumezanu_Cheng_Ni_Zong_Chen_Chawla_2019} & 0.70 & 16.2 & 30.87 & no & no \\
    \hline
    Vanilla LSTM \cite{filonov2016multivariateindustrialtimeseries} & 0.67 & 15.42 & 36.02 & no & no \\
    \hline
    Conv-AE \cite{conv_ae}	& 0.66 & 5.58 & 46.05 & no & no \\
    \hline
    LSTM-AE \cite{lstm_ae} & 0.65 & 14.59 & 39.42 & no & no \\
    \hline
    \textbf{AAD-LLM} & \textbf{0.56} & \textbf{47.6} & \textbf{31.7} & \textbf{yes} & \textbf{yes} \\
    \hline
    LSTM-VAE \cite{Bowman2015GeneratingSF} & 0.56 & 9.2 & 54.81 & no & no \\
    \hline
    Vanilla AE \cite{Chen2017OutlierDW} & 0.45 & 7.55 & 66.57 & no & no \\
    \hline
    Isolation forest \cite{4781136} & 0.4 & 6.86 & 72.09 & no & no \\
    \hline
    Null detector & 0 & 100 & 100 & & \\
    \hline
    \end{tabular}
    \caption{Best outlier detection scores for each anomaly detection method implemented on the SKAB dataset, sorted by F1 score \cite{Elhalwagy_2022}. A selection of NNs and ML based fault detection methods were chosen to compare on the benchmarks. RAAD-LLM and RAAD-LLMv2 metrics are averaged over the best 5 model runs. Multimodality allows for the enriching of input series data with semantics to enable more collaborative decision-making between the model and plant operators. For this work, multimodality refers to a model being optimized to detect anomalies across both time-series data and text. A model that requires no training or fine-tuning on the data it is applied to is considered transferable with zero-shot capabilities. Unlike all other methods, AAD-LLM, RAAD-LLM, and RAAD-LLMv2 are not trained or fine-tuned on the dataset they are applied to and are multimodal without requiring any additional strategies.} 
    \label{table:tab_results2}
    \end{center}
\end{table*}

To assess the performance of the frameworks, we applied RAAD-LLM and RAAD-LLMv2 to both the SKAB and use-case datasets. Evaluation metrics include accuracy, precision, recall, and F1-score, with a particular focus on the model's ability to reduce false positives and improve anomaly detection rates when compared to the original AAD-LLM. 

The brief results are shown in \autoref{table:tab_results1}. With $95\%$ confidence, for the use-case dataset, RAAD-LLM achieved an accuracy of $88.6 \pm 2.1\%$, which is a significant improvement over the baseline model. Furthermore, RAAD-LLM's precision of $92.6 \pm 0.1\%$, recall of $91.1 \pm 3.3\%$ and F1 score of $91.9 \pm 1.7\%$ are all notable improvements over the previous architecture. For the SKAB dataset, RAAD-LLM achieved an accuracy of $71.6 \pm 0.4\%$, F1 score of $73.5 \pm 0.8\%$, FAR of $42.1 \pm 0.9\%$, and MAR of $11.4 \pm2.3\%$. As with the use-case dataset, all evaluation metrics for the SKAB dataset show a significant improvement over the previous architecture.


While RAAD-LLMv2 injects only relevant information and eliminates the need for manual context restructuring, it exhibited lower performance metrics for both datasets when compared to RAAD-LLM. This performance trade-off highlights the challenges posed by dynamic knowledge retrieval in the RAAD-LLMv2 framework. Although RAAD-LLM is shown to be highly effective in controlled scenarios where manual context restructuring is feasible, RAAD-LLMv2 is a more scalable alternative for real-world scenarios requiring automated domain knowledge retrieval. These findings reveal opportunities for further optimization to improve RAAD-LLMv2's overall performance.


\autoref{table:tab_results2} summarizes the scores for algorithms on $3$ application benchmarks using the SKAB dataset, sorted by F1 score. For F1 score, bigger is better. For both FAR and MAR, less is better. While our previous architecture ranked $8^{th}$ among all NN and ML based methods, RAAD-LLM and RAAD-LLMv2 ranked $1^{st}$ and $4^{th}$, respectively in F1 score. Although, both RAAD-LLM and RAAD-LLMv2 ranked last in FAR, they ranked $1^{st}$ and $2^{nd}$, respectively in MAR. In industrial applications where there is potential for severe safety implications and the risk of catastrophic failure, the MAR is generally considered more important and is often prioritized. Effective anomaly detection systems should strive to minimize both FAR and MAR, but special attention should be given to ensuring that real anomalies are not overlooked, as the consequences of such oversights can far outweigh the inconveniences posed by false alarms.

The integration of the RAG component into the RAAD-LLM and RAAD-LLMv2 frameworks has led to marked improvements in anomaly detection performance compared to the previous AAD-LLM architecture. Results indicate that RAG enhances the model's performance in detecting anomalies within time series data. Our findings affirm the efficacy of RAG in augmenting the capabilities of LLMs in PdM applications. With RAAD-LLM outperforming all presented fault detection methods, repurposing LLMs with RAG integration is shown effective in detecting anomalies in time series data accurately. Overall, our findings support the use of LLMs for anomaly detection for the PdM use-case, underlining their capability and potential in handling challenges in time series anomaly detection in data-constrained industrial applications. This work significantly advances anomaly detection methodologies, potentially leading to a paradigm shift in how anomaly detection is implemented across various industries.

%% file: component_subject_sections/Conclusion/conclusion.tex
In conclusion, the RAAD-LLM framework demonstrates significant advancements in anomaly detection by leveraging the integration of the RAG pipeline, multimodal capabilities, and zero-shot transferability to address the challenges of data-sparse industrial environments. By accessing external knowledge bases and enriching data inputs, the model enhances interpretability and has been shown to be superior to baseline methods in identifying and classifying anomalies. Furthermore, RAAD-LLM’s emphasis on minimizing MAR ensures its suitability for safety-critical industrial applications.

Despite these achievements, areas for improvement remain. Future work should prioritize automating domain context restructuring to reduce the reliance on manual intervention, which can be time-intensive. RAAD-LLMv2 was designed to address this issue and be more scalable in real-world settings. However, it exhibited slightly lower performance compared to RAAD-LLM. Fine-tuning LlamaIndex configurations or developing hybrid approaches that blend manual context restructuring with automated retrieval should be explored. Additionally, RAAD-LLM was applied to only static datasets to better understand how processes failed after the failure had already occurred. Transitioning from static datasets to real-time data streams would expand RAAD-LLM’s applicability to online anomaly detection. This would enable more proactive and dynamic monitoring systems. Lastly, further exploration into extending the methodology beyond sensor data to other domains could broaden the impact of this framework across diverse industries. This methodology could be extended to areas such as financial fraud detection (transaction data) or healthcare diagnostics (image and medical data). This extension could involve reconfiguring the RAG process or adaptability mechanism to handle these new data types and scenarios.

Ultimately, RAAD-LLM represents a promising shift in how anomaly detection is approached, balancing interpretability, accuracy, and adaptability to meet the growing demands of modern industrial applications.